\documentclass[letterpaper, 10 pt, conference]{ieeeconf}  %

\IEEEoverridecommandlockouts                              %

\usepackage[space]{cite}
\usepackage{amsmath,amssymb,amsfonts}
\usepackage{graphicx}
\usepackage{booktabs}
\usepackage{multirow}
\usepackage{algorithm}
\usepackage{algpseudocode}
\usepackage[hidelinks]{hyperref}
\usepackage{pifont}%
\usepackage[subtle,tracking=normal,wordspacing=normal]{savetrees}

\usepackage[font=footnotesize]{caption}
\allowdisplaybreaks

\newcommand{\diag}{\operatorname{diag}}
\newcommand{\clip}{\operatorname{clip}}
\newcommand{\norm}[1]{\left\lVert #1 \right\rVert}

\newcommand{\T}{^\top}

\newcommand{\xmark}{\ding{55}}%

\title{\LARGE \bf
ElastiQP: An Always-Feasible QP Solver \\ for Constrained Robot Control
}

\author{Daniel Morton \quad Jon Arrizabalaga \quad Zachary Manchester \quad Marco Pavone%
\thanks{Daniel Morton was supported by a NASA Space Technology Graduate Research Opportunity}%
\thanks{Daniel Morton and Marco Pavone are with the Departments of Mechanical Engineering and Aeronautics \& Astronautics, Stanford University, USA. Jon Arrizabalaga and Zachary Manchester are with the Department of Aeronautics and Astronautics, Massachusetts Institute of Technology, USA. Correspondence: {\tt\small dmorton@stanford.edu}}%
}

\begin{document}

\maketitle
\thispagestyle{empty}
\pagestyle{empty}

\begin{abstract}

As robot capabilities increase, quadratic programming (QP)-based controllers must account for a similarly increasing number of constraints to ensure safe, reliable operation. Yet, with each added constraint, this introduces more chances of \textit{momentary conflict}: in which case, a QP solver that returns an ``infeasible'' status leaves the controller with nothing to execute. To address this, we introduce ElastiQP, a modified dual active-set QP solver that relaxes every inequality constraint with an exact, per-constraint \(\ell_1\) penalty while keeping equality constraints (dynamics) hard. Notably, ElastiQP does so by folding the slack variables into the solver \textit{analytically}, maintaining a constant size of the condensed linear system. On a suite of robot control benchmarks, ElastiQP achieves microsecond-level performance, matching or outperforming leading modern solvers on feasible problems. On infeasible problems, ElastiQP handles these gracefully, confining violations to strictly the conflicting inequality terms, returning a usable solution up to 40x faster than the best alternative solvers. ElastiQP is available as an open-source C++ header-only library, with Python and JAX interfaces, at 
\mbox{{\small\url{https://github.com/StanfordASL/elastiqp}}}.

\end{abstract}

\section{Introduction}

Quadratic programs (QPs) are a central part of model-based robot control, including whole-body control \cite{kuindersma2016atlas}, safety filters \cite{ames2017cbf, morton2025oscbf, morton2026constrained}, hierarchical control \cite{escande2014hierarchical}, legged locomotion \cite{dicarlo2018cheetah}, and inverse kinematic and dynamic methods more generally.
Broadly speaking, QP-based controllers can be divided into two main categories: small, dense problems for real-time (single-timestep, high-frequency) control, and larger, more structured problems that consider a predictive horizon (e.g., model predictive control (MPC)). 
For real-time control, the target rate for such controllers is typically 1~kHz, to stabilize high-frequency dynamic modes. This necessitates a solver which can reliably deliver a solution in (ideally well under) 1~ms, for problems on the order of tens of decision variables.

The constraints in these QPs typically encode dynamics terms (equalities), or limits, tasks, and safety margins (usually, inequalities). For complex robots and environments, these constraints can easily number in the hundreds, often introducing edge cases where these are \textit{instantaneously} inconsistent, particularly under unexpected disturbances.
In this case, a solver that returns ``infeasible'' leaves the robot with no control to execute, a potentially dangerous state. Ideally, even in the case of conflict, the controller will always return a minimally-violating, best-effort command within the time budget.

\begin{figure}[t]
    \centering
    \includegraphics[width=\linewidth]{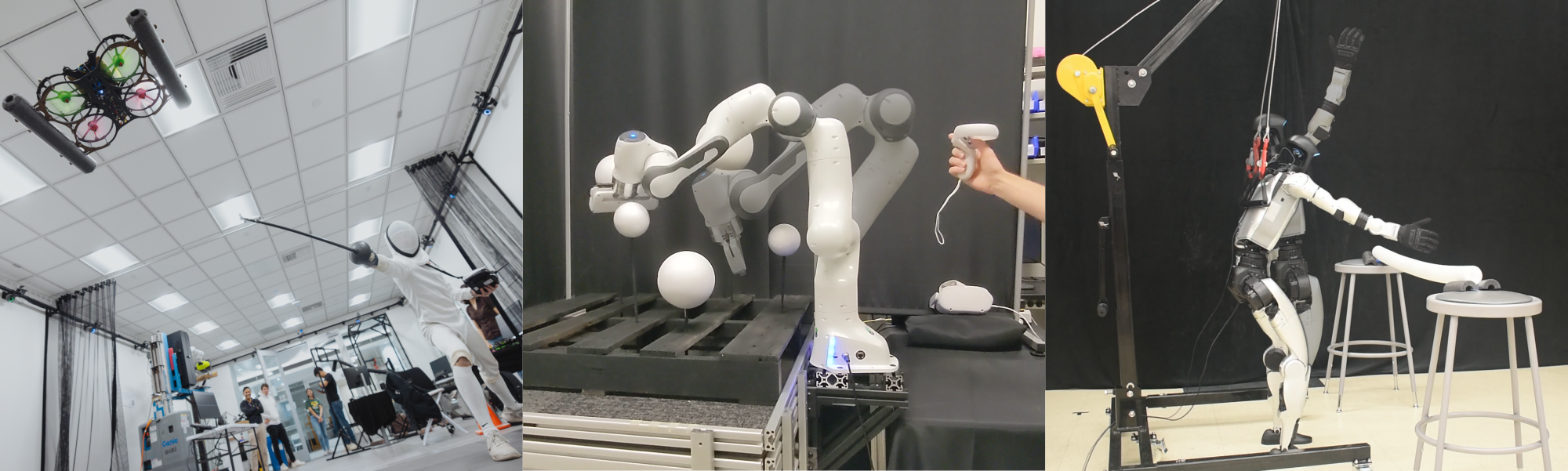}
    \caption{Elastic QPs are at the core of hardware deployments of online safety filters and controllers, including on drones, manipulators \cite{morton2025oscbf}, and humanoids \cite{morton2026constrained}. In each case, elasticity resolves momentary conflict between constraints (safety, actuator limits, and tasks) in a minimally-conservative manner, allowing the robot to operate reliably near its dynamic limits. ElastiQP now accelerates these same problems, for high-frequency deployment on even the most challenging highly-constrained and high-DoF systems.}
    \label{fig:hardware}
    \vspace{-3mm}
\end{figure}

A formulation well-suited to this setting is an \textit{elastic} QP: each constraint is relaxed with an exact \(\ell_1\) penalty on violation (the ``elastic mode'' of SNOPT \cite{gill2002snopt}). 
When the problem is feasible, and if elastic penalty terms are sufficiently high, the elastic solution is \textit{exactly} the hard solution. Thus, for the majority of timesteps when no conflict occurs, the added reliability of the elastic structure costs nothing to the optimal control. When conflict does occur, the \(\ell_1\) structure encourages sparsity in the constraint violation, restricting the relaxation to only the terms in conflict. Conversely, an \(\ell_2\) relaxation spreads constraint violation across terms which were never in conflict, leading to undesired tradeoffs in these edge cases. 

\textit{Remark}: For robot control, we are interested in a specific type of elastic mode, where we allow for both hard equality constraints, and elastic inequality constraints. Dynamics constraints are posed as equalities, and these are always physically consistent with each other, and meaningless if relaxed (for instance, a solution where torques and accelerations are incompatible). If equalities are possibly-inconsistent, non-dynamics terms, these can always be made elastic via opposing elastic inequalities.

In general, the only way to pose an elastic QP to a general-purpose solver is to add a slack variable per constraint and solve over the expanded decision vector. With large numbers of constraints, the factorization cost is dominated by the slack variables, rather than the original decision variables, leading to difficulty meeting the 1~ms time budget for control. 
This is the case for ProxQP \cite{bambade2023proxsuite}, PIQP \cite{schwan2023piqp}, and DAQP \cite{arnstrom2022daqp}, three of the leading solvers for problems at this scale. Alternatively, ProxQP and DAQP each offer a ``closest feasible" or ``soft" mode in the case of primal infeasibility, but these solve for the (undesirable) minimum \(\ell_2\) shift in the constraints, require first detecting infeasibility, and cannot individually weight relaxation on a per-constraint basis. ProxQP and PIQP's sparse backends can also exploit the (considerable) sparsity that per-constraint slack variables introduces, but sparse backends are primarily designed for large-scale problems, and perform poorly in this small-scale high-frequency setting.
qpax \cite{tracy2024differentiability, arrizabalaga2026differentiableinteriorpointmethodsingle} performs competitively here: its elastic mode introduced a trick for slack variable elimination in interior point methods, which inspired this work. However, this elastic mode does not support equality constraints or warm-starting, its speed falls short of 1~kHz rates on humanoid-scale problems, and it tends to fail to achieve tight tolerance on even moderately ill-conditioned problems. 
FlexQP \cite{oshin2026deep} adopts a similar \(\ell_1\) relaxation on \textit{both} inequality and equality constraints, within an OSQP-style ADMM method \cite{stellato2020osqp}. This work targets hyperparameter learning and batched solves on GPU at loose tolerances, rather than tight, warm-started solves at control rates. FlexQP also converges at first-order rates, and iterates over the slacks and duals, rather than eliminating these in closed form.

\subsection{Statement of Contributions}

We present ElastiQP, an always-feasible QP solver for constrained robot control. ElastiQP is a modified dual active-set method, designed for hard equality constraints, \(\ell_1\) elastic inequality constraints, and strong warm-starting performance across the repetitive structure of control loops. Notably, ElastiQP solves these problems through an exact elimination of the elastic slacks, significantly reducing the computational cost of solving an \(\ell_1\)-relaxed QP as compared to carrying these as decision variables. In dual active-set methods, this reduces to a simple \([0, w]\) box constraint on the dual variables, with a corresponding update to the working-set solve to account for the saturated terms. We also provide an open-source, header-only C++/Eigen implementation of ElastiQP, along with Python and JAX interfaces, at 
\mbox{{\small\url{https://github.com/StanfordASL/elastiqp}}}. %

\subsection{Paper Organization}

In Section \ref{sec:problem}, we define the elastic QP and its key properties. Section \ref{sec:elastic_condensation} then introduces the elastic solver, and characterizes the differences from traditional active-set methods. In Section \ref{sec:results}, we then study where conflicts arise in robotics problems, and how the solution to an elastic QP is well-suited to balance conflicts. Additionally, we extensively benchmark ElastiQP against other leading solvers on relevant feasible and infeasible robotics scenarios, and in Section \ref{sec:maros}, on a broader suite of challenging problems.

\section{The Elastic Quadratic Program}
\label{sec:problem}

\begin{figure*}
    \centering
    \includegraphics[width=0.85\linewidth]{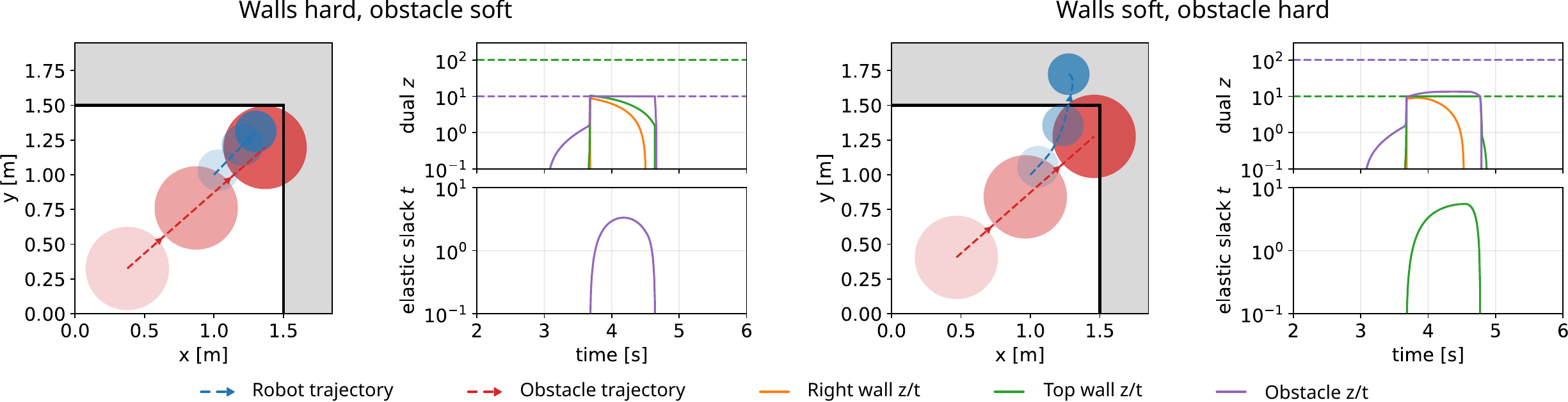}
    \caption{\textbf{When conflict between constraints is inevitable, what solution should be returned?} Consider a 2D double-integrator robot which has been backed into a corner, with a dynamic obstacle moving towards it. Inherently, this leads to a conflict between three safety constraints: stay within the two boundaries of the workspace, and avoid collision with moving obstacle. In edge cases like this, we would prefer a solver which returns a \textit{reasonable} balance between conflicts, rather than returning an infeasible status. Here, the balance can be set via the magnitude of the penalty terms, \(w\). For \(w_{\text{wall}} > w_{\text{obs}}\) (left), the optimal action is to accept collision with the obstacle while avoiding the walls, whereas for \(w_{\text{wall}} < w_{\text{obs}}\), the robot accepts violating the workspace boundary to avoid collision with the dynamic obstacle. In either case, when conflict occurs, the optimal dual \(z\) for the relaxed constraint reaches its corresponding cap \(w\), allowing the elastic slack \(t\) to grow when the conflict is active.}
    \label{fig:constraint_conflict}
    \vspace{-3mm}
\end{figure*}

ElastiQP solves the \textit{elastic} QP
\begin{equation}
\label{eq:elastic_qp}
\begin{aligned}
\min_{x,\,t} \quad & \tfrac{1}{2}x^\top Qx + q^\top x + w^\top t \\
\text{s.t.} \quad & Ax = b \\
& Gx - t \leq h \\
& t \geq 0
\end{aligned}
\end{equation}
with decision variable \(x \in \mathbb{R}^n\), elastic slacks \(t \in \mathbb{R}^p\), cost data \(Q \in \mathbb{S}_+^n\), \(q \in \mathbb{R}^n\), hard equality constraints \(A \in \mathbb{R}^{m\times n}\), \(b \in \mathbb{R}^m\), elastic inequality constraints \(G \in \mathbb{R}^{p\times n}\), \(h \in \mathbb{R}^p\), and penalty weights \(w \in \mathbb{R}^p_{>0}\).
At any \(x\), the inner minimization over \(t \geq 0\) is solved by \(t_i = \max(0, [Gx-h]_i)\), so \eqref{eq:elastic_qp} is equivalent to the exact \(\ell_1\) penalty formulation
\begin{equation}
\label{eq:l1_form}
\begin{aligned}
\min_{x} \quad & \tfrac{1}{2}x^\top Qx + q^\top x
  + \textstyle\sum_i w_i \max(0,\, [Gx - h]_i) \\
\text{s.t.} \quad & Ax = b
\end{aligned}
\end{equation}
and \(t\) recovers the per-row constraint violations. Due to the elastic slacks, the inequality constraints can never cause infeasibility, and the problem \eqref{eq:elastic_qp} is feasible if and only if \(Ax = b\) is consistent.

Introducing multipliers \(y \in \mathbb{R}^m\) for \(Ax = b\), and \(z_t, z \in \mathbb{R}^p\) for \(t \geq 0\) and \(Gx - t \leq h\) respectively, stationarity in \(t\) gives \(w - z_t - z = 0\), and yields two key properties \cite{nocedal2006numerical, gill2002snopt}:
\begin{enumerate}
\item \textbf{Bounded multipliers.} At any optimum, \(0 \leq z \leq w\) elementwise, with \(z_t = w - z\). The bound is the dual expression of the \(\ell_1\) penalty: swapping the infinite wall of a hard constraint for a ramp of slope \(w_i\) clips the corresponding multiplier from \([0, \infty)\) to \([0, w_i]\). \label{bounded_multipliers}
\item \textbf{Exactness.} If the hard-constrained QP (with \(t \equiv 0\) enforced) is feasible with an optimal inequality multiplier \(z^\star\) satisfying \(z^\star_i < w_i\) for all \(i\), then every solution of \eqref{eq:elastic_qp} has \(t = 0\) and its \(x\) solves the hard-constrained QP \cite[Thm.~17.3]{nocedal2006numerical}. \label{exactness}
\end{enumerate}
Exactness is what distinguishes this relaxation from generic soft constraints, which trade constraint satisfaction against the objective on every solve. 
When elastic constraints are compatible (and when the penalties \(w\) are set greater than the optimal duals \(z\)), the relaxation has no impact on the optimal solution, and activates only under conflict. 

\subsection[Relaxation: Why l1 and not l2]{Relaxation: Why \(\ell_1\) and not \(\ell_2\)}
\label{sec:l1_vs_l2}

When constraints do conflict, \(\ell_1\) and \(\ell_2\) relaxations differ considerably in how violations are distributed. The \(\ell_1\) penalty is exact, with a sparse violation structure: only the constraints that cannot be jointly satisfied are relaxed, while all other constraints hold exactly. Whereas, a quadratic \(\ell_2\) penalty is inexact and diffuse, spreading small violations across many constraints, even those that were not in conflict in the first place. Even on feasible problems, an \(\ell_2\) relaxation can lead to small constraint violations, as the cost to violate these vanishes at zero.

Consider, for instance, a robot controller where the inequality terms encode both task and actuator limits. If the robot is operating at high speeds, a task constraint may be inconsistent with actuator limits (the robot cannot physically produce a torque that satisfies the task). In this setting, the desirable behavior is for \textit{only} the task to be relaxed. With an \(\ell_1\) structure, setting the per-constraint penalty values \(w_i\) according to the desired relaxation hierarchy \(w_{\text{task}} < w_{\text{actuator}}\) provides exactly this outcome. We further discuss and quantify the difference between an \(\ell_2\)-relaxed baseline in Section \ref{sec:infeasible_control_loops}, and provide a visual example of constraint conflict in Fig. \ref{fig:constraint_conflict}.

\subsection{Equality constraints: Hard versus elastic}

ElastiQP considers elastic inequalities with \textit{hard equalities}, specifically to match the typical structure of robot control QPs. In these problems, equality terms typically encode dynamics and are consistent by construction, so the hardness assumption here retains the persistent feasibility benefits of the elastic inequalities. An alternative structure would be to similarly relax the dynamics equalities, encoding them as elastic rows with large penalties. However, this introduces additional concerns, the main being that the guarantee that the returned control is \textit{dynamically consistent} is lost, despite these constraints never being the sole source of conflict. In our experiments, we find that by keeping these constraints hard, even when inequalities are in conflict, these terms hold to tight tolerance (Table \ref{tab:violation_structure}). 

If, however, equality terms correspond to softer tasks which may be relaxed in the case of inconsistency, ElastiQP does support elastic equalities, via pairs of opposing elastic inequality constraints \(\beta \leq a\T x \leq \beta\) with corresponding penalty terms \(w_E\). In general, this structure should be reserved only for the case where equalities may be soft, due to the additional cost imposed by adding two inequalities for every equality.

\section{The Elastic Condensation}
\label{sec:elastic_condensation}

Before noting the adjustments required to make a reliable elastic solver, consider first a hard-constrained QP (equivalent to \eqref{eq:elastic_qp} with the elastic slacks \(t\) fixed at 0):
\begin{equation}
\label{eq:standard_qp}
\begin{aligned}
\min_{x} \quad & \tfrac{1}{2}x\T Qx + q\T x \\
\text{s.t.} \quad & Ax = b \\
& Gx \leq h
\end{aligned}
\end{equation}

A general-purpose QP solver which targets this problem structure can always solve the elastic problem \eqref{eq:elastic_qp} by appending the elastic slacks \(t\) to the primal variables \(x\) as a new, expanded decision vector. This creates a problem with \(n + p\) unknowns and \(2p\) inequality constraints, leading to the most expensive part of the solver (factorizations) similarly growing as \(p\) increases. In robot control, \(p\) is typically several times \(n\), and as such, adding elasticity in this expanded form significantly increases computational cost. 

ElastiQP instead \textit{condenses} the problem: \(t\) is eliminated analytically, maintaining a constant \(n \times n\) factorization size and strong solver performance even as \(p \gg n\). Critically, we find that \textit{this condensation strategy is not unique to any one solver method}: similar condensation strategies can be efficiently implemented across the three main families of QP solvers (interior point (IPM), augmented Lagrangian (ALM), and active set methods (ASM)). 

In short, the relationship between condensation strategies across solver families stems from Property \ref{bounded_multipliers}: in each case, the inequality multipliers \(z\) are restricted to \([0, w]\) as opposed to \([0, \infty)\). Whereas each hard-constrained solver (implicitly or explicitly) classifies an inequality as \textit{inactive} or \textit{active}, a condensed elastic solver introduces a new state, \textit{saturated}. 
\begin{table}[ht]
\centering
\caption{Inequality States for Elastic QPs}
\label{tab:inequality_states}
\begin{tabular}{lll}
\toprule
State & Multiplier & Inequality \\
\midrule
Inactive  & \(z_i = 0\)          & Satisfied, \([Gx - h]_i < 0\), \(t_i = 0\) \\
Active    & \(0 < z_i < w_i\)    & Tight, \([Gx - h]_i = 0\), \(t_i = 0\) \\
Saturated & \(z_i = w_i\)        & Violated, \(t_i = [Gx - h]_i > 0\) \\
\bottomrule
\end{tabular}
\vspace{-2mm}
\end{table}

Saturation is the only state where the elastic slacks \(t\) are nonzero: the constraint is violated, resulting in a cost of \(w_i\) per unit of violation (see: \eqref{eq:l1_form}). 

If, then, an efficient elastic condensation can be posed for each solver family, which of IPM/ALM/ASM is best-suited to the robot control domain? We find that \textit{active-set methods} consistently outperform IPM and ALM for these problems, and as such, in the following subsection, we focus our discussion on the elastic condensation strategy for ASMs. However, we include the condensations for IPMs and ALMs in the Appendix, sections \ref{sec:ipm_condensation} and \ref{sec:alm_condensation}: these strategies form the basis of our ElastiQP-IPM and ElastiQP-PDAL variants, which we evaluate in Section \ref{sec:results}.

\subsection{Preliminaries: Active-Set Methods}

At the solution of \eqref{eq:standard_qp}, some inequality constraints are tight (holding with equality), and the rest are inactive. If the set of tight constraints, the \textit{active set}, were known in advance, the inequalities could be dropped or converted to equalities, and the QP would reduce to an equality-constrained QP: one fast linear solve. An active-set method \cite{nocedal2006numerical} searches for this set. These methods maintain a \textit{working set}, \(\mathcal{W}\) (an estimate of the active set), solve the equality-constrained QP in which the constraints in \(\mathcal{W}\) hold with equality, and use the result to revise \(\mathcal{W}\). A constraint is added to \(\mathcal{W}\) when the current iterate violates it, and removed when its multiplier is negative. 

This method has good properties for robot control, particularly when considering warm-starting. Because each iteration changes \(\mathcal{W}\) by one row, the factorization can be updated rather than fully recomputed, and the cost of a solve is proportional to the number of changes to the working set. When warm-starting from a previous solution in consecutive timesteps, the active set is generally quite stable, and the iteration count for the warm solve is only the number of constraints whose status (active/inactive) has changed. In control, this is often zero or one. Conversely, this same property can lead to long iteration times on a cold-start, where the number of updates to \(\mathcal{W}\) is as large as the number of initial active constraints.

A \textit{primal} active-set method \cite{nocedal2006numerical} maintains a primal-feasible iterate, which requires a feasible starting point, and iteratively adds a constraint to \(\mathcal{W}\) when a step would otherwise violate it. A \textit{dual} active-set method \cite{goldfarb1983numerically, arnstrom2022daqp} maintains a dual-feasible iterate instead, satisfying primal feasibility only at termination. 
Compared to primal methods, a dual ASM does not require a primal-feasible starting point, so when warm starting, the previous working set and multipliers can be reused without first restoring feasibility under the new data. Standard dual methods do require \(Q \succ 0\) so that \(x\) can be eliminated in closed-form, but this can be resolved with proximal point iterations to handle the semidefinite case \cite{bemporad2018prox}.

Dual ASMs operate on the dual problem,
\begin{equation}
\label{eq:dual_problem}
\begin{aligned}
\max_{\lambda = [y;\, z]} \quad & -\tfrac12 \norm{M\T \lambda}^2 - d\T \lambda \\
\text{s.t.} \quad & z \geq 0
\end{aligned}
\end{equation}

Here, we introduce the following intermediate terms: \(C = [A;\, G]\) and \(c = [b;\, h]\) are the stacked constraint terms, \(R\) is a Cholesky factor of \(Q\) (upper triangular with \(R^\top R = Q\)),  \(\lambda = [y;\, z]\) are the multipliers, and we have \(M = C R^{-1}\), \(v = R^{-\top} q\) and \(d = c + M v\), similarly to \cite{arnstrom2022daqp}. A solution to the primal problem \eqref{eq:standard_qp} can be recovered from the dual optimum as
\begin{equation}
    \label{eq:primal_recovery}
    x(\lambda) = -R^{-1}(M^\top \lambda + v)
\end{equation}
Though \eqref{eq:dual_problem} is itself a QP, the constraints (\(z \geq 0\)) are simple to handle for an active-set method. 

A subproblem of a dual ASM restricts \eqref{eq:dual_problem} to \(\mathcal{W}\): this fixes \(\lambda_i = 0\) for \(i \notin \mathcal{W}\), so that \(M\T\lambda = M_{\mathcal{W}}\T\lambda_{\mathcal{W}}\) and \(d\T\lambda = d_{\mathcal{W}}\T\lambda_{\mathcal{W}}\) (denoting \(M_{\mathcal{W}}\), \(d_{\mathcal{W}}\), and \(\lambda_{\mathcal{W}}\) as the rows of \(M\), \(d\), and \(\lambda\) indexed by \(\mathcal{W}\)). This yields an unconstrained problem
\begin{equation}
\label{eq:subproblem}
\max_{\lambda_{\mathcal{W}}} \quad -\tfrac12 \norm{M_{\mathcal{W}}\T \lambda_{\mathcal{W}}}^2 - d_{\mathcal{W}}\T \lambda_{\mathcal{W}}
\end{equation}
whose maximizer is the solution of the linear system
\begin{equation}
\label{eq:subproblem-solve}
\big(M_{\mathcal{W}} M_{\mathcal{W}}\T\big)\, \lambda_{\mathcal{W}} = -d_{\mathcal{W}}
\end{equation}

A full iteration of the dual ASM is, therefore:
\begin{enumerate}
\item  Compute the multipliers \(\lambda^\star_{\mathcal{W}}\) that maximize the dual over the working set, from \eqref{eq:subproblem-solve}.
\item If moving from the current multipliers to \(\lambda^\star_{\mathcal{W}}\) would push one of them negative, stop at the first to reach zero, remove that index from \(\mathcal{W}\), and go back to step 1.
\item  Otherwise accept \(\lambda^\star_{\mathcal{W}}\), evaluate the constraints at \(x(\lambda^\star)\), and add the most violated row to \(\mathcal{W}\). If none is violated, that point is optimal.
\end{enumerate}

In the hard-constrained case, if the dual is unbounded, the primal is infeasible.

\subsection{An Elastic Condensation for Dual Active-Set Methods}
\label{sec:asm_condensation}

An efficient implementation of the \(\ell_1\)-relaxed elastic form for dual ASMs is a \textit{single change} to the dual problem \eqref{eq:dual_problem}: in addition to the non-negativity bound on the inequality multipliers \(z\), we simply add an upper bound at the penalty values, \(w\). \eqref{eq:dual_problem} then becomes
\begin{equation}
\label{eq:elastic_dual_problem}
\begin{aligned}
\max_{\lambda = [y;\, z]} \quad & -\tfrac12 \norm{M\T \lambda}^2 - d\T \lambda \\
\text{s.t.} \quad & 0 \leq z \leq w
\end{aligned}
\end{equation}
directly following from Property \ref{bounded_multipliers} (bounded multipliers).

To solve \eqref{eq:elastic_dual_problem}, this requires very small changes to the iteration, mainly, that we introduce a \textit{saturated} set \(\mathcal{S}\) containing the inequality constraint indices where \(\lambda_i = w_i\). With this, the unconstrained problem of \eqref{eq:subproblem} then becomes 
\begin{equation}
\label{eq:elastic-subproblem}
\max_{\lambda_{\mathcal{W}}} \quad -\tfrac12 \norm{M_{\mathcal{W}}\T \lambda_{\mathcal{W}} + M_{\mathcal{S}}\T w_{\mathcal{S}}}^2 - d_{\mathcal{W}}\T \lambda_{\mathcal{W}}
\end{equation}
whose maximizer is the solution of the elastic linear system,
\begin{equation}
\label{eq:elastic-subproblem-solve}
\big(M_{\mathcal{W}} M_{\mathcal{W}}\T\big)\, \lambda_{\mathcal{W}} = -d_{\mathcal{W}} - M_{\mathcal{W}} M_{\mathcal{S}}\T w_{\mathcal{S}}
\end{equation}

An iteration of the elastic dual ASM is very similar to the non-elastic case:
\begin{enumerate}
\item Compute the multipliers \(\lambda^\star_{\mathcal{W}}\) that maximize the dual over the working set, from \eqref{eq:elastic-subproblem-solve}.
\item If moving from the current multipliers to \(\lambda^\star_{\mathcal{W}}\) would push one of them out of the \([0, w_i]\) box, stop at the first to reach the bound, remove that index from \(\mathcal{W}\), add it to \(\mathcal{S}\) if it reached \(w_i\), and go back to step 1.
\item  Otherwise accept \(\lambda^\star_{\mathcal{W}}\) and evaluate the constraints at \(x(\lambda^\star)\). An index outside \(\mathcal{W}\) is wrong if the corresponding constraint is inactive but violated, or saturated but strictly satisfied; add the constraint index that is wrong by the largest amount to \(\mathcal{W}\). If no index is wrong, that point is optimal.
\end{enumerate}
In short, in step 1 we replace \eqref{eq:subproblem-solve} with \eqref{eq:elastic-subproblem-solve}, and in steps 2 and 3 we account for the saturated set \(\mathcal{S}\) and the box bounds in the working set update. 

In the elastic case, the dual can only become unbounded through the equality multipliers \(y\), and thus the elastic dual ASM will \textit{always} return a solution whenever the equality constraints are consistent (as they are in robot dynamics). 

Hence, given the natural incorporation of elasticity into dual ASMs, and the strong performance of DAQP \cite{arnstrom2022daqp} on feasible robot problems, we build ElastiQP's core solver methods using DAQP as our primary reference. In the following section, we evaluate this choice against not only DAQP, but also equivalent strategies and solvers within the IPM and ALM families, to exhaustively validate and identify the best method for elastic robot control.

\section{Results: Robot Control}
\label{sec:results}

\begin{table}[t]
\centering
\smallskip
\caption{Robot Control Problem Structures \\ ($n$ variables, $m$ equalities, $p$ inequalities).}
\label{tab:scenarios}
\footnotesize
\setlength{\tabcolsep}{4pt}
\begin{tabular}{llcccc}
\toprule
Scenario & Robot & \(x\) & \(n\) & \(m\) & \(p\) \\
\midrule
arm-osc & 6-DoF arm & $\tau$ & 6 & 0 & 24 \\
biman-ik & Two 6-DoF arms, rigid grasp & $[\dot q_1; \dot q_2]$ & 12 & 6 & 48 \\
hum-wbc & 28-DoF humanoid, 2 foot contacts & $[\ddot q;\, f]$ & 46 & 18 & 132 \\
\bottomrule
\end{tabular}
\vspace{-3mm}
\end{table}

\begin{table*}[t]
\centering
\smallskip
\caption{Average Solve Times in \(\mu s\) (cold/warm) for Feasible and Infeasible Control Loops (\(\epsilon_{\text{abs}}=10^{-6}\))}
\label{tab:robot}
\footnotesize
\setlength{\tabcolsep}{9.0pt} %
\begin{tabular}{lllccccccc}
\toprule
& & & \multicolumn{3}{c}{Feasible control loops} & \multicolumn{3}{c}{Infeasible control loops} \\
\cmidrule(lr){4-6} \cmidrule(lr){7-9}
Solver & Family & Form & arm-osc & biman-ik & hum-wbc & arm-osc & biman-ik & hum-wbc \\
\midrule
ElastiQP & ASM & $\ell_1$ & 1.3 / 1.3 & 4.0 / 3.7 & 52 / 45 & \textbf{2.1} / \textbf{1.3} & \textbf{5.9} / \textbf{3.9} & \textbf{178} / \textbf{54} \\
ElastiQP-PDAL & ALM & $\ell_1$ & 2.7 / 1.8 & 7.0 / 5.7 & 115 / 55 & 7.2 / 3.4 & 40 / 19 & 463 / 116  \\
ElastiQP-IPM & IPM & $\ell_1$ & 18 & 45 & 932 & 16 & 50 & 947 \\
DAQP & ASM & hard & \textbf{0.5} / \textbf{0.4} & \textbf{3.0} / \textbf{2.9} & \textbf{44} / \textbf{43} & \xmark\space(1.2 / 1.2) & \xmark\space(3.3 / 3.4) & \xmark\space(76 / 76) \\
DAQP-soft & ASM & $\ell_2$ & 0.5 / 0.4 & 3.3 / 3.3 & 55 / 52 & 1.7 / 0.9 & 4.3 / 3.6 & 104 / 57 \\
DAQP-slack & ASM & $\ell_1$ & 23 / 15 & 154 / 100 & 3249 / 2066 & 24 / 14 & 169 / 117 & 4175 / 2329 \\
ProxQP & ALM & hard & 11 / 8.0 & 25 / 25 & 180 / 79 & \xmark\space(20 / 8268) & \xmark\space(53 / 15062) & \xmark\space(799 / 96137) \\
ProxQP-closest & ALM & $\ell_2$ & 13 & 29 & 263 & 74187\(^\dagger\) & 2521\(^\dagger\) & 36512\(^\dagger\) \\
ProxQP-slack & ALM & $\ell_1$ & 235 / 21 & 1343 / 88 & 30783 / 1224 & 180 / 24 & 1506 / 121 & 34445 / 2587 \\
ProxQP-sparse & ALM & $\ell_1$ & 215 / 18 & 919 / 45 & 25800 / 454 & 198 / 57 & 1169 / 77 & 29242 / 1024 \\
PIQP & IPM & hard & 17 & 44 & 552 & \xmark\space(77) & \xmark\space(253) & \xmark\space(1421) \\
PIQP-slack & IPM & $\ell_1$ & 94 & 408 & 8839 & 77 & 453 & 8289 \\
PIQP-sparse & IPM & $\ell_1$ & 47 & 147 & 4756\(^\dagger\) & 42 & 164 & 4368 \\
qpax\(^*\) & IPM & hard & 63 & 148 & 659 & \xmark\space(848) & \xmark\space(3189) & \xmark\space(13403) \\
qpax-elastic\(^*\) & IPM & $\ell_1$ & 105 & 160 & 1602 & 95 & 542\(^\dagger\) & 16215\(^\dagger\)  \\
\bottomrule
\addlinespace
\multicolumn{9}{l}{\footnotesize \xmark \space  No usable solution at any timestep; times reported are the average time to termination.} \\
\multicolumn{9}{l}{\footnotesize \(\dagger\) Convergence or certification failures occurred on some but not all timesteps.} \\
\multicolumn{9}{l}{\footnotesize \(*\) qpax struggles to converge at \(\epsilon_{\text{abs}}=10^{-6}\). For these tests only, we run at qpax's (looser) default tolerance of \(\epsilon_{\text{abs}}=10^{-5}\).}
\end{tabular}
\vspace{-3mm}
\end{table*}

\begin{table}[ht]
\centering
\smallskip
\caption{Violation structure for inequality and equality constraints on infeasible controls (hum-wbc)}
\label{tab:violation_structure}
\begin{tabular}{@{}lccc@{}}
\toprule
Solver                    & Violated Rows & $\norm{[Gx -h]_+}_1$ & $\norm{Ax-b}_\infty$ \\ \midrule
ElastiQP (\(\ell_1\))     & 1    & 2.51     & \(10^{-8}\)        \\
DAQP-soft (\(\ell_2\))    & 28   & 147      & \(10^{-11}\)      \\
ProxQP-closest (\(\ell_2\))  & 29   & 3.10     & \(0.2\)        \\ \bottomrule
\end{tabular}
\vspace{-3mm}
\end{table}

ElastiQP is designed with constrained robot control in mind, and as such, this is primary focus area for our experiments. We consider three representative robotics applications of increasing difficulty (Table \ref{tab:scenarios}): task-space inverse dynamics of a single 6-degree-of-freedom (DoF) arm (arm-osc), differential inverse kinematics of a 12-DoF bimanual system (biman-ik), and whole-body control of a 28-DoF humanoid robot (hum-wbc). Together, these span the range of typical problem sizes and structures seen in short-horizon robot control.

\subsection{Problem Setup}

Of the three problems, the simplest is the task-space inverse dynamics of a single 6-DoF arm. Here, we solve for the joint torques to achieve a given end-effector acceleration, with upper/lower limits on the joint velocities and torques. 
Next, we consider bimanual differential inverse kinematics with two 6-DoF arms, with an equality constraint between the end-effector motions mimicking a rigid grasp constraint, and upper/lower limits on the joint positions and velocities. This more than doubles the problem size compared to the single arm, and adds hard equality constraints to better evaluate the mixed hard/elastic setting. 
Our most challenging setting is humanoid whole-body control: for a 28-DoF humanoid robot with a floating root, we optimize over the generalized accelerations \(\ddot{q}\) and the contact wrenches \(f\) for each foot, assuming planar double-support contact. The equality constraints enforce the underactuated dynamics and contacts, and the inequalities handle the upper/lower limits on the joint velocities and torques, as well as the linearized wrench cones for the feet.  

Offline, for each problem, we pre-solve a 500-timestep closed-loop trajectory using Pinocchio \cite{carpentier2019pinocchio} for the kinematics and dynamics terms, and cache the resulting QP matrices at each timestep. Then, during benchmarking, we replay the cached QP sequence, allowing us to isolate the timing of the QP solve from the kinematics and dynamics computation, while maintaining a representative slowly-drifting control-loop setting. We also set the actuator limits such that for each problem, approximately 50\% of the timesteps are constrained at the optimum, to ensure that both unconstrained and constrained states are represented in the timing. When constructing the infeasible timesteps, we take the feasible QP and tighten a single inequality constraint past what is admissible from the other constraints -- this also allows us to clearly observe if a solution violated more than the one necessary constraint.

\subsection{Evaluation}

For comparison, we evaluate ElastiQP against three leading solvers for the small, dense problem setting: DAQP \cite{arnstrom2022daqp}, PIQP \cite{schwan2023piqp} and ProxQP \cite{bambade2023proxsuite}, as well as qpax \cite{tracy2024differentiability, arrizabalaga2026differentiableinteriorpointmethodsingle}, which implements an efficient condensed elastic mode similar to ElastiQP. Additionally, to properly evaluate \textit{``which family of solvers is best-suited to elastic robot control problems?"}, we compare ElastiQP's active set method against our own elastic PDAL and IPM variants (further detailed in the Appendix). 

Across each of these solvers, we also consider five different formulations of the problem. \textit{Elastic} indicates an \(\ell_1\)-relaxed inequality constraint structure with analytical condensation of the elastic slack variables: the main strategy of ElastiQP and of qpax's elastic form. \textit{Hard} solves the standard hard-constrained problem, with no additional protection against infeasibility. \textit{Slack} indicates an expanded \(\ell_1\)-relaxed form where both \(x\) and \(t\) are stacked into the decision vector, for persistent feasibility even when solved with a hard-constrained routine. \textit{Closest} refers to ProxQP's closest-feasible mode, an \(\ell_2\)-relaxed problem form similar to DAQP's \textit{soft} form. Finally, \textit{sparse} indicates solving the expanded \(\ell_1\) form with a sparse matrix backend (to potentially take advantage of the sparsity of elastic slacks). For solvers that support warm-starting (ElastiQP, DAQP, and ProxQP), we report timing values for both cold and warm solves.

All results were recorded on a laptop with an Intel Core Ultra 7 258V CPU, and (for C++ solvers) compiled with \texttt{gcc} 13.3 and \texttt{-O3} optimization. Note that compiling with \texttt{-march=native} tends to give an additional \(\sim1.5\times\) performance gain on larger-scale problems, so the reported numbers are conservative. qpax is a purely JAX/Python library, and values reflect the JIT-compiled CPU performance using a recent JAX version (0.11.0), and double precision.

\subsection{Feasible Control Loops}

From Table \ref{tab:robot}, ElastiQP demonstrates extremely fast (microsecond-level) performance on both feasible and infeasible domains, with even cold-started solves on the hardest infeasible problems (hum-wbc) falling comfortably below the 1~ms target required for 1~kHz control. Warm-starting reliably delivers an additional 2-4x in performance improvement on larger-scale problems, making ElastiQP the most reliable, always-feasible method for delivering high-frequency solves on challenging control problems.

In the strictly-feasible setting, DAQP is the fastest method for solving these problems, with ElastiQP following closely behind. Active set methods (such as DAQP or ElastiQP) perform very strongly in this setting on both cold and warm starts -- there are relatively few constraints active at the optimum (fast cold starts) and the active set is relatively stable across timesteps (fast warm starts). The additional handing for elasticity costs only a few microseconds at these problem scales; an acceptable trade for the reliability of an always-feasible method. The PDAL variants (ProxQP, ElastiQP-PDAL) also exhibit good warm-started performance, but fall behind the active-set variants, particularly on the smaller-scale problems.  

Compared to the expanded slack form of DAQP, we can clearly see the benefits of ElastiQP's condensation (an improvement of over 40x). For ElastiQP's PDAL and IPM variants, the condensation improves performance by 10x and 20x, respectively, with this gap increasing to over 250x when looking at cold-started PDAL. ElastiQP is also 30-80x faster than qpax, our main elastic-condensed baseline. 

\subsection{Infeasible Control Loops}
\label{sec:infeasible_control_loops}

In the infeasible setting, ElastiQP dramatically outperforms all alternative solvers and strategies for posing the relaxed problem. As expected, standard hard-constrained solvers (DAQP, PIQP, ProxQP, and qpax's hard form) fail to solve an infeasible problem, and we note that reporting infeasibility can sometimes take a (dangerously) long time. Consider a warm-started controller, using ProxQP: if even a single timestep is momentarily infeasible, it could take upwards of 93 milliseconds to report infeasibility and manage this, compared to the nominal 1 millisecond budget. 

ProxQP's \(\ell_2\)-relaxed ``closest-feasible" mode does report a solution, but not in a reasonable amount of time for control. Additionally, this \(\ell_2\) mode exhibits the undesirable relaxation structure, as discussed in Section \ref{sec:l1_vs_l2} and seen in Table \ref{tab:violation_structure}. This mode results in small violations to 29 constraints when only 1 was in conflict, \textit{and} no longer holds the equality constraints to tight tolerances, whereas all other \(\ell_1\)-based methods successfully achieve a minimally-relaxed problem. DAQP's soft \(\ell_2\) returns a solution much faster than ProxQP's closest-feasible mode, and keeps equalities consistent, but at the cost of significantly higher \(\ell_1\) violation than necessary.

The second-best solver to ElastiQP in this setting is ProxQP's sparse backend, solving the expanded slack form, but only when warm-started. Even still, this method is not fast enough for reliable 1~kHz control on humanoid-scale problems, and is around 20-40x slower than ElastiQP. The other alternatives, ProxQP-slack and PIQP-slack, fall further behind in speed. 

qpax is unreliable at \(10^{-6}\) precision, frequently failing to converge on these problems. Even after reducing the tolerance to \(10^{-5}\), we still observe convergence failures on the harder biman-ik and hum-wbc problems, when constraint conflict appears. While qpax's elastic backend has previously demonstrated good performance on smaller, single-arm problems (see \cite{morton2025oscbf}), it should not be relied on for larger control problems that may be momentarily infeasible.

\section{Results: Maros--M\'esz\'aros}
\label{sec:maros}

\begin{figure*}[ht]
    \centering
    \smallskip
    \includegraphics[width=0.9\linewidth]{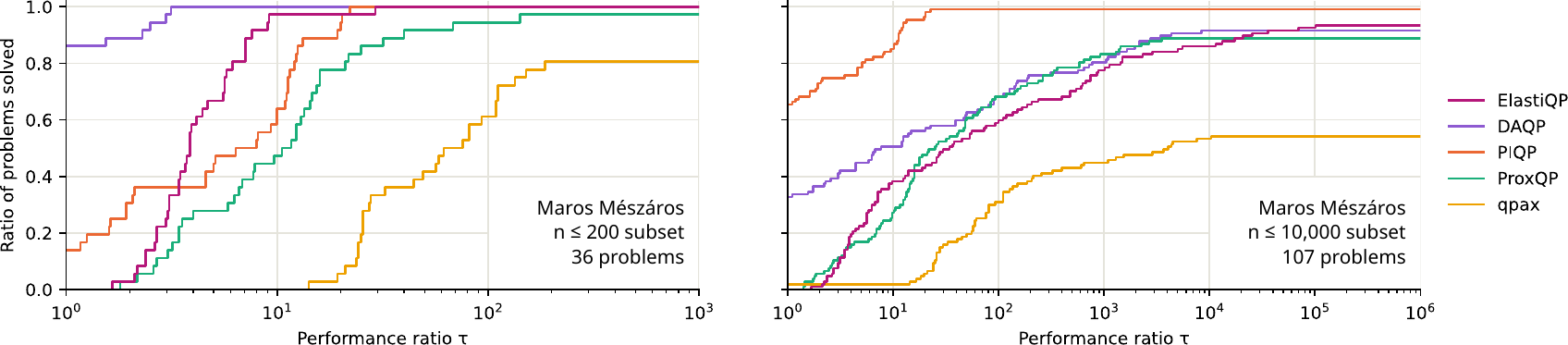}
    \caption{Maros--M\'esz\'aros performance profiles: 36-problem \(n \leq 200\) subset (left) and 107-problem \(n \leq 10,000\) subset (right), \(\epsilon_{\text{abs}}=10^{-6}\). In settings \textit{well outside of the robotics domain}, ElastiQP is competitive with the best general-purpose solvers, in both speed and reliability, even on the ill-conditioned, large, and sparse problems of this set.}
    \label{fig:maros}
    \vspace{-3mm}
\end{figure*}

While robot control is ElastiQP's primary focus, the Maros--M\'esz\'aros test set provides a complementary benchmark on poorly-scaled, challenging problems \cite{maros1999repository}. 
The test set contains a wide range of problem sizes and sparsity, many of which are out of scope for a solver targeted towards small, dense problems. Despite this, we find it insightful to evaluate ElastiQP on these problems, to identify any potential robustness issues,  to determine if the \(\ell_1\) relaxation has led to noticeable performance reduction, and to identify trends between solver families (IPM, ALM, ASM). 
We divide our analysis into two subsets: the \(n \leq 200\) subset allows us to compare performance on small-to-medium-sized ill-conditioned problems, while the \(n \leq 10,000\) subset pushes the limits of our dense solver. 
As these problems are known to be feasible, for elastic solvers (ElastiQP, qpax), we set the penalties above the optimal dual required for the problem so that by exactness (Property \ref{exactness}) the elastic and hard solutions coincide.

As seen in Figure. \ref{fig:maros}, we see that (as anticipated) ElastiQP is \textit{not} the best solver for these problems: DAQP wins on the smaller-scale set, while PIQP comfortably wins at larger scales. However, this test does reveal key properties of ElastiQP: most importantly, that \textit{the elastic condensation does not degrade the solver's performance for general-purpose problems}. In both the small and large subsets, ElastiQP performs comparably to the best solvers, and ElastiQP even solves some problems that DAQP and ProxQP do not -- for instance, on the \(n \leq 200\) subset, ProxQP reports a (premature) infeasible status on the QRECIPE problem. The only baseline that implements the condensed elastic form, qpax, trails significantly behind ElastiQP in both speed and reliability.

\section{Additional Design Notes}

When working with ElastiQP (and \(\ell_1\) relaxations more broadly), the value of the elastic penalty, \(w\), must be set by the user prior to solving the problem. By exactness (Property \ref{exactness}), if the problem is feasible and the penalty is greater than the optimal dual of the hard-constrained problem, then the hard solution will be recovered. This is the desired behavior for control: on feasible timesteps, the elastic relaxation should be effectively invisible. How then, should \(w\) be set without \textit{a priori} knowledge of the solution?

\begin{table}[ht]
\centering
\smallskip
\caption{Maximum Duals for Benchmark Problems}
\label{tab:max_duals}
\footnotesize
\begin{tabular}{@{}llll@{}}
\toprule
Problem set                   & \% of set  & Examples                   & \(\| z \|_{\infty}\)      \\ \midrule
Robot control                 & 100\%    & arm-osc, biman-ik, hum-wbc & \(\leq 1.0\) \\
Maros--M\'esz\'aros & 22\%    & DUAL4, HS21, QAFIRO        & \(\leq 1.0\) \\
Maros--M\'esz\'aros & 50\% & CVXQP2\_S, LOTSCHD         & \(\leq 1e3\) \\
Maros--M\'esz\'aros & 89\%   & DUALC2, QRECIPE            & \(\leq 1e6\) \\
\bottomrule
\end{tabular}
\end{table}

In general, this will be problem-dependent, but the robot control problems from our benchmarks (Section \ref{sec:results}) typically have very small duals, often \(\leq 1\) (Table \ref{tab:max_duals}). Empirically, we find that around \(w=1e3\) is a reasonable default for control -- striking a balance between being sufficiently high to ensure there is no unnecessary relaxation, but not too high that it interferes with the numerical conditioning of the problem. Other problem domains may require higher penalties: within the Maros subset from Section \ref{sec:maros}, 50\% of the problems would be exact with a \(1e3\) penalty, 89\% with \(1e6\), and the remainder requiring higher values. Very few problems (e.g. QPCBOEI2) had an optimal dual above \(1e8\); above this value is where conditioning becomes a concern.

\section{Conclusion}
\label{sec:conclusion}

In this work, we have presented ElastiQP, a QP solver designed for robot control, with the implicit reliability of elastic, \(\ell_1\)-relaxed inequality constraints for times where persistent feasibility of the controller cannot be guaranteed. ElastiQP is intended for immediate integration into existing controllers and safety layers for robotics, providing reliable, fast solve times even when compute is limited, or when high-DoF robots necessitate larger, highly-constrained QPs. Initial work towards differentiability is in progress, with GPU acceleration for batched QP solves as the next step. In future work, we envision ElastiQP applied as a fast, elastic inner solver for SQP pipelines, particularly for global inverse kinematics and retargeting. The condensation can also be expanded to a block-sparse form, for incorporating elastic constraints into highly-structured MPC and trajectory optimization problems.

\bibliographystyle{IEEEtran}
\bibliography{references}

\begin{thebibliography}{10}
\providecommand{\url}[1]{#1}
\csname url@samestyle\endcsname
\providecommand{\newblock}{\relax}
\providecommand{\bibinfo}[2]{#2}
\providecommand{\BIBentrySTDinterwordspacing}{\spaceskip=0pt\relax}
\providecommand{\BIBentryALTinterwordstretchfactor}{4}
\providecommand{\BIBentryALTinterwordspacing}{\spaceskip=\fontdimen2\font plus
\BIBentryALTinterwordstretchfactor\fontdimen3\font minus \fontdimen4\font\relax}
\providecommand{\BIBforeignlanguage}[2]{{%
\expandafter\ifx\csname l@#1\endcsname\relax
\typeout{** WARNING: IEEEtran.bst: No hyphenation pattern has been}%
\typeout{** loaded for the language `#1'. Using the pattern for}%
\typeout{** the default language instead.}%
\else
\language=\csname l@#1\endcsname
\fi
#2}}
\providecommand{\BIBdecl}{\relax}
\BIBdecl

\bibitem{kuindersma2016atlas}
S.~Kuindersma \emph{et~al.}, ``Optimization-based locomotion planning, estimation, and control design for the atlas humanoid robot,'' \emph{Autonomous Robots}, vol.~40, no.~3, pp. 429--455, Mar 2016.

\bibitem{ames2017cbf}
A.~D. Ames, X.~Xu, J.~W. Grizzle, and P.~Tabuada, ``Control barrier function based quadratic programs for safety critical systems,'' \emph{IEEE TAC}, vol.~62, no.~8, pp. 3861--3876, 2017.

\bibitem{morton2025oscbf}
D.~Morton and M.~Pavone, ``Safe, task-consistent manipulation with operational space control barrier functions,'' in \emph{IEEE/RSJ IROS}, 2025.

\bibitem{morton2026constrained}
D.~Morton, P.~Mohnot, and M.~Pavone, ``Constrained whole-body tracking for humanoid robots,'' \emph{arXiv:2606.00374}, 2026.

\bibitem{escande2014hierarchical}
A.~Escande, N.~Mansard, and P.-B. Wieber, ``Hierarchical quadratic programming: Fast online humanoid-robot motion generation,'' \emph{IJRR}, vol.~33, no.~7, p. 1006–1028, Jun. 2014.

\bibitem{dicarlo2018cheetah}
J.~Di~Carlo \emph{et~al.}, ``Dynamic locomotion in the mit cheetah 3 through convex model-predictive control,'' in \emph{IEEE/RSJ IROS}, 2018.

\bibitem{gill2002snopt}
P.~E. Gill, W.~Murray, and M.~A. Saunders, ``{SNOPT}: An {SQP} algorithm for large-scale constrained optimization,'' \emph{SIOPT}, vol.~12, no.~4, pp. 979--1006, 2002.

\bibitem{bambade2023proxsuite}
A.~Bambade \emph{et~al.}, ``{ProxQP}: an efficient and versatile quadratic programming solver for real-time robotics applications and beyond,'' \emph{IEEE T-RO}, pp. 1--19, 2025.

\bibitem{schwan2023piqp}
R.~Schwan, Y.~Jiang, D.~Kuhn, and C.~N. Jones, ``{PIQP}: A proximal interior-point quadratic programming solver,'' in \emph{IEEE CDC}, 2023.

\bibitem{arnstrom2022daqp}
D.~Arnström, A.~Bemporad, and D.~Axehill, ``A dual active-set solver for embedded quadratic programming using recursive ${LDL}^{T}$ updates,'' \emph{IEEE TAC}, vol.~67, no.~8, pp. 4362--4369, 2022.

\bibitem{tracy2024differentiability}
K.~Tracy and Z.~Manchester, ``On the differentiability of the primal-dual interior-point method,'' \emph{arXiv:2406.11749}, 2024.

\bibitem{arrizabalaga2026differentiableinteriorpointmethodsingle}
J.~Arrizabalaga, K.~Tracy, and Z.~Manchester, ``A differentiable interior-point method in single precision,'' \emph{arXiv:2605.17913}, 2026.

\bibitem{oshin2026deep}
A.~Oshin, R.~V. Ghosh, A.~D. Saravanos, and E.~Theodorou, ``Deep flex{QP}: Accelerated nonlinear programming via deep unfolding,'' in \emph{ICLR}, 2026.

\bibitem{stellato2020osqp}
B.~Stellato, G.~Banjac, P.~Goulart, A.~Bemporad, and S.~Boyd, ``{OSQP}: an operator splitting solver for quadratic programs,'' \emph{Mathematical Programming Computation}, vol.~12, no.~4, pp. 637--672, 2020.

\bibitem{nocedal2006numerical}
J.~Nocedal and S.~J. Wright, \emph{Numerical Optimization}, 2nd~ed.\hskip 1em plus 0.5em minus 0.4em\relax Springer, 2006.

\bibitem{goldfarb1983numerically}
D.~Goldfarb and A.~Idnani, ``A numerically stable dual method for solving strictly convex quadratic programs,'' \emph{Mathematical programming}, vol.~27, no.~1, pp. 1--33, 1983.

\bibitem{bemporad2018prox}
A.~Bemporad, ``A numerically stable solver for positive semidefinite quadratic programs based on nonnegative least squares,'' \emph{IEEE TAC}, vol.~63, no.~2, pp. 525--531, 2018.

\bibitem{carpentier2019pinocchio}
J.~Carpentier \emph{et~al.}, ``The pinocchio c++ library -- a fast and flexible implementation of rigid body dynamics algorithms and their analytical derivatives,'' in \emph{IEEE SII}, 2019.

\bibitem{maros1999repository}
I.~Maros and C.~Mészáros, ``A repository of convex quadratic programming problems,'' \emph{Optimization Methods and Software}, vol.~11, no. 1-4, pp. 671--681, 1999.

\bibitem{conn1991globally}
A.~R. Conn, N.~I.~M. Gould, and P.~Toint, ``A globally convergent augmented lagrangian algorithm for optimization with general constraints and simple bounds,'' \emph{SINUM}, vol.~28, no.~2, pp. 545--572, 1991.

\end{thebibliography}

\appendix

In this appendix, we discuss how elasticity can be incorporated into other families of solvers (IPM, ALM): these alternative strategies were benchmarked in Section \ref{sec:results} to identify the best-suited solver for problems in robotics.

\subsection{An Elastic Condensation for IPMs}
\label{sec:ipm_condensation}

Within the broad class of IPMs, PIQP \cite{schwan2023piqp} demonstrates excellent performance across a wide range of problems (including the small, dense problems of robot control), and we focus on this solver when discussing IPMs. PIQP differs from other IPMs in that it wraps the interior point iteration in a proximal method of multipliers %
with primal and dual proximal weights \(\rho, \delta\) for regularization. Given this, PIQP's KKT matrix \(K\) can be written as
\begin{equation}
\label{eq:piqp-K}
K = Q + \rho I + \tfrac{1}{\delta} A\T A + G\T \diag(1/W)\, G
\end{equation}
with \(W = s/z + \delta \). Notably, the weight \(1/W_i\) is large when an inequality constraint is tight, and vanishes when the constraint is inactive.

In the expanded \((x, t)\) form of the elastic QP, each inequality yields two constraint rows, one for \(Gx - t \leq h\) (with slack \(s_{in}\) and multiplier \(z_{in}\)) and another for \(t \geq 0\) (with slack \(s_t\) and multiplier \(z_t\)). Here, we have weights \(1/W_{in}\) and \(1/W_t\) where \(W_{in} = s_{in}/z_{in} + \delta\) and \(W_t = s_t/z_t + \delta\). The penalty terms \(w\) appear in the optimality conditions only through stationarity in \(t\): \(w - z_{in} - z_t = 0\).

Noticing that \(t\) enters the Newton system only through diagonal terms, \(dt\) can be eliminated exactly as \(ds\) and \(dz\) were. This yields \eqref{eq:piqp-K} with the weights \(\diag(1/W)\) replaced with \(\diag(\Lambda)\), where
\begin{equation}
\label{eq:m-ipm-K}
\Lambda_i = \frac{\gamma_i}{W_{in,i}} \qquad
\gamma_i = \frac{\rho + 1/W_{t,i}}{\rho + 1/W_{t,i} + 1/W_{in,i}} \in (0, 1)
\end{equation}
The factor \(1/W_{in}\) is the hard-constraint weight, and \(\gamma_i\) is the elastic correction, which defines the \textit{saturated} state:
\begin{itemize}
\item An \textit{active} row has \(t_i \to 0\)  and \(\gamma_i \approx 1\): the row is weighted exactly as a hard constraint.
\item A \textit{saturated} row has \(t_i > 0\) and \(\Lambda_i \to \rho / (1 + \rho W_{in,i}) \approx \rho\): the row drops out of \(K\) up to the proximal regularization. 
\item An \textit{inactive} row has \(1/W_{in,i} \to 0\), and drops out of \(K\) regardless of \(\gamma_i\), as in the hard method.
\end{itemize}

With no regularization, \(\rho = \delta = 0\), and \(\Lambda_i\) reduces to \(1 / (s_{in} / z_{in} + s_t / z_t)_i\), the same weight as qpax's elastic form~\cite{tracy2024differentiability}.

\subsection{An Elastic Condensation for ALMs}
\label{sec:alm_condensation}

For \textit{primal-dual} ALMs such as ProxQP \cite{bambade2023proxsuite}, the elastic condensation reduces down to a \textit{single} modification: everywhere a hard-constrained solver projects the multiplier estimate onto \(z \geq 0\) with \(\max(0, \cdot)\), the condensed form replaces this with a clamp onto the box \([0, w]\), 
\begin{equation}
\label{eq:m-zhat}
\hat z_i = \clip_{[0,\, w_i]}\big(\tilde z_i\big)
\end{equation}
with \(\tilde z(x) = z^k + (Gx - h) / \mu_{in}\) representing the unclamped multiplier estimate at the \(k^{th}\) step, and \(\mu_{in}\) being the AL penalty parameter for inequality constraints.

The AL penalty term for an inequality constraint therefore becomes a Huber-like function: quadratic near the boundary and then linear once the constraint is in conflict: 
\begin{equation}
\label{eq:psi}
\psi_i(\tilde z_i) = \mu_{in} \cdot
  \begin{cases}
      0 & \tilde z_i \leq 0 \\[2pt]
      \tfrac12 \tilde z_i^2 & 0 < \tilde z_i < w_i \\[4pt]
      w_i \tilde z_i - \tfrac12 w_i^2 & \tilde z_i \geq w_i 
  \end{cases}
\end{equation}
This function is \(C^1\) and piecewise quadratic, with kinks in the gradient at \(\tilde z_i = 0\) and \(\tilde z_i = w_i\), so a semismooth Newton method still applies. The only elastic consequences for the Newton method are (1) handling saturated constraints, whose multipliers are set to \(w_i\) and are excluded from the system, and (2) considering two breakpoints per constraint in the exact line search rather than one.

These elastic changes to the inner loop are minimal, but they change the meaning of the primal residuals, which has downstream implications for the outer loop. In the always-feasible elastic setting, the residuals measure ``how far are the multipliers from their final values", rather than ``how far is \(x\) from feasibility". \(1/\mu_{in}\) also acts as a dual step length, and this must be sufficiently large to allow the multiplier \(z_i\) to reach its final value in a reasonable amount of iterations. Under a standard bound-constrained Lagrangian (BCL) schedule \cite{conn1991globally, nocedal2006numerical}, we observed occasional convergence issues in the elastic PDAL outer loop, particularly when warm-starting and immediately entering or departing regions where constraints were in conflict (see: the jumps in the optimal duals in Figure \ref{fig:constraint_conflict}). 

Resolving this problem required several updates to the BCL \(\mu\) schedule, but the most critical include (1) avoiding cold-resets of \(\mu\), which can reset any progress made in allowing \(z_i\) to climb or descend towards its limits at 0 or \(w_i\); and (2) when immediately entering or leaving conflict regions, instead of shrinking \(\mu\) by a fixed factor, jump directly to the known target \(\mu\) which is computable from the inactive and saturated multiplier targets 0 or \(w_i\). For brevity, we leave a full discussion of these (heuristic) strategies to a future expanded paper.

\end{document}